\pdfoutput=1

\documentclass[11pt]{article}

\usepackage[]{acl}

\usepackage{times}
\usepackage{latexsym}

\usepackage[T1]{fontenc}

\usepackage[utf8]{inputenc}

\usepackage{microtype}

\usepackage{inconsolata}

\usepackage{amsmath}
\usepackage{booktabs}
\usepackage{multirow}
\usepackage{graphicx}
\usepackage{xspace}
\usepackage[export]{adjustbox}

\newcommand{\method}[1]{\textsc{#1}\xspace}
\newcommand{\MarginContrast}{\method{MarginContrast}}
\newcommand{\PairContrast}{\method{PairContrast}}
\newcommand{\MLE}{\method{MLE}}
\newcommand{\SeqDistill}{\method{SeqDistill}}
\newcommand{\SWING}{\method{Swing}}
\newcommand{\FACTPEGASUS}{\method{FactPegasus}}
\newcommand{\HUMANREF}{\method{HumanRef}}

\newcommand{\dataset}[1]{#1\xspace}
\newcommand{\SAMSUM}{\dataset{SAMSum}}
\newcommand{\DIALOGSUM}{\dataset{DialogSum}}

\newcommand{\UniEval}{\method{UniEval}}
\newcommand{\AlignScore}{\method{AlignScore}}
\newcommand{\GEval}{\method{G-Eval}}

\title{Factual Dialogue Summarization via Learning from Large Language Models}

\author{Rongxin Zhu \quad Jey Han Lau \quad Jianzhong Qi \\
        School of Computing and Information Systems \\ 
        The University of Melbourne \\ 
        \texttt{rongxinz1@student.unimelb.edu.au, \{laujh, jianzhong.qi\}@unimelb.edu.au}}

\begin{document}
\maketitle
\begin{abstract}
Factual consistency is an important quality in dialogue summarization. Large language model (LLM)-based automatic text summarization models generate more factually consistent summaries compared to those by smaller pretrained language models, but they face deployment challenges in real-world applications due to privacy or resource constraints. In this paper, we investigate the use of symbolic knowledge distillation  to improve the factual consistency of smaller pretrained models for dialogue summarization. We employ zero-shot learning to extract symbolic knowledge from LLMs, generating both factually consistent (positive) and inconsistent (negative) summaries. We then apply two contrastive learning objectives on these summaries to enhance smaller summarization models. Experiments with BART, PEGASUS, and Flan-T5 indicate that our approach surpasses strong baselines that rely on complex data augmentation strategies. Our approach achieves better factual consistency while maintaining coherence, fluency, and relevance, as confirmed by various automatic evaluation metrics. We also provide access to the data and code to facilitate future research~\footnote{\url{https://github.com/731935354/symbolic_distill_contrastive_summ}}.

\end{abstract}

\section{Introduction}
Automatic text summarization aims to create a concise summary of a source document that keeps all the essential points. Although current models are capable of generating fluent and coherent summaries, one main issue is factual inconsistency, where generated summaries are found to contain facts that are absent from or contradict the source~\citep{maynez2020faithfulness, huang2021factual}.
To tackle this, a number of methods have been proposed, including explicit fact modeling~\cite{zhu2021enhancing, huang2020knowledge}, post-editing~\cite{lee2022factual, balachandran2022correcting, chen2021improving} and contrastive learning~\cite{wan2022factpegasus, cao2021cliff, liu2021co2sum}. Contrastive learning-based methods, in particular,  offer a straightforward solution without requiring any modification to the model architecture, but their performance hinges on careful and often rule-based construction of negative samples~\citep{cao2021cliff, liu2021co2sum, wan2022factpegasus}.

\begin{figure}[t]
\centering
\includegraphics[width=\linewidth]{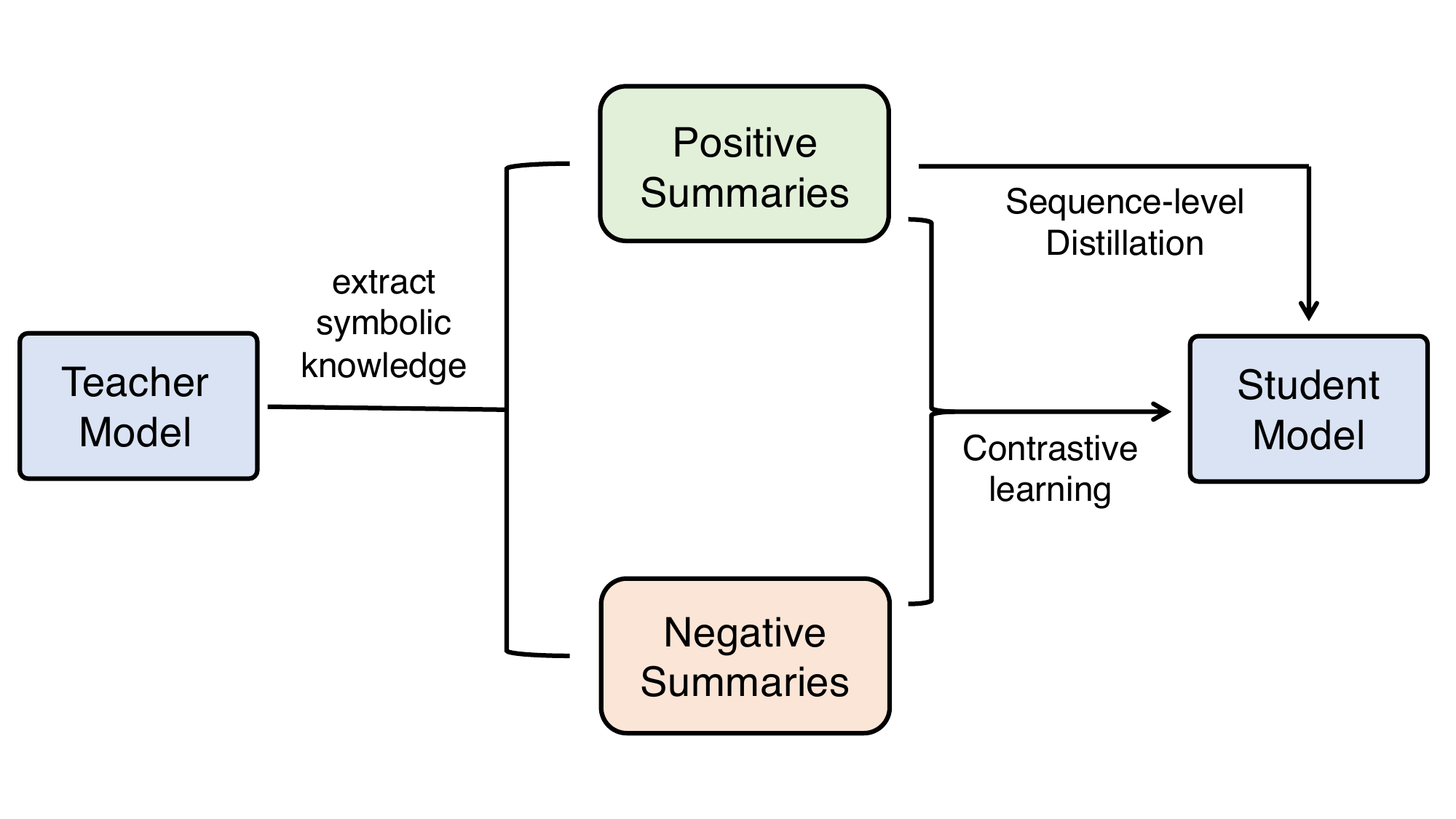}
\caption{An overview of our framework to leverage symbolic knowledge distillation to improve the factual consistency for smaller (student) models in dialogue summarization.}
\label{fig:outline}
\end{figure}

The rise of large language models (LLMs) changed the landscape of NLP, and they exhibit emergent capabilities~\cite{wei2022emergent} such as in-context learning~\cite{brown2020language, min2022rethinking} and instruction following~\cite{ouyang2022training}. We have seen zero- or few-shot prompting with LLMs achieving strong performance on various NLP tasks~\citep{wei2021finetuned, ye2021crossfit} including summarization~\citep{zhang2023benchmarking}, showing better coherence, relevance and factual consistency than human-written reference summaries.

Although impressive, LLMs are not always deployable in real-world applications due to substantial computational resources~\citep{strubell2019energy} or privacy concerns (as many state-of-the-art LLMs are closed source and can only be accessed via APIs). Thus, it is important to construct more cost-efficient and compact models with similar summarization capabilities. To this end, knowledge distillation~\citep{hinton2015distilling} --- a technique that can transfer the knowledge from a large \textit{teacher model} to a small \textit{student model} --- has been explored~\cite{sun2020mobilebert, aguilar2020knowledge}. Symbolic knowledge distillation~\citep{west2022symbolic}, a special form of knowledge distillation, extracts symbolic knowledge (e.g., textual information) from the teacher model and uses such knowledge as training signal for the student model. This method is especially useful when working with blackbox teacher models where we do not have access to their output probability distribution (which is the case for closed source LLMs such as ChatGPT).

In this paper, we explore symbolic knowledge distillation to improve the factual consistency of (smaller) pretrained models in dialogue summarization. Concretely, we extract symbolic knowledge from an LLM teacher (\textit{gpt-3.5 turbo}) in the format of \textbf{positive summaries} and \textbf{negative summaries}. Positive summaries are factually consistent with the source article (i.e., a dialogue) while negative summaries are not. We experiment with various strategies to incorporate these summaries and train the student model, including sequence-level knowledge distillation~\cite{kim2016sequence} and two contrastive learning-based methods. Our experiments cover three widely used pretrained models: BART~\cite{lewis2020bart}, PEGASUS~\cite{zhang2020pegasus}, and Flan-T5~\cite{chung2024scaling} on two popular dialogue summarization datasets: SAMSum~\citep{gliwa2019samsum} and DialogSum~\cite{chen2021dialogsum}.

To summarize, our contributions are as follows:

\begin{itemize}
    \item We propose to improve the factual consistency of (small) dialogue summarization models via symbolic knowledge distillation from LLMs.
    
    \item We experiment with LLMs to generate not only factually consistent summaries but also inconsistent ones, and we incorporate such summaries to train small dialogue summarization models with two contrastive objectives.
    
    \item We discovered that: (1) symbolic knowledge distillation enables us to create smaller dialogue summarization models that surpass strong baselines; and (2) the top-performing student model achieves comparable or even better factual consistency compared to human-written references without compromising other quality dimensions such as fluency or coherence.
\end{itemize}

\section{Related Work}
\subsection{Evaluating and Enhancing Factual Consistency}
We summarize two areas of factuality research:  \emph{evaluation} and \emph{enhancement}.

Automatic evaluation metrics are generally constructed on question-answering systems~\citep{fabbri-etal-2022-qafacteval, scialom-etal-2021-questeval, durmus-etal-2020-feqa, manakul2023mqag} or textual entailment models~\citep{kryscinski-etal-2020-evaluating, goyal-durrett-2020-evaluating, laban2022summac, Zhang2024FineGrainedNL}. More recent methods leverage the capability of LLMs to follow zero-shot and few-shot instructions~\citep{fu2023gptscore, min2023factscore, liu2023gpteval}. Another line of work aims at developing metrics that can detect the factual consistency between text pairs in different tasks~\citep{deng2021compression, zha-etal-2023-alignscore}, such as a knowledge-grounded dialogue.

Methods to enhance the factual consistency of summarization models mainly fall into the following categories: explicit modeling of the facts in source documents~\cite{zhu2021enhancing, huang2020knowledge}, post-editing model generated summaries for better factual consistency~\cite{lee2022factual, balachandran2022correcting, chen2021improving}, training summarization model with less noisy data by data filtering~\citep{nan2021entity, goyal2021annotating, wan2022factpegasus}, and data augmentation-based methods~\citep{wang2022improving, adams2022learning}. The last category is usually combined with contrastive learning~\citep{wan-bansal-2022-factpegasus, liu2021co2sum, cao2021cliff}, which has shown a high effectiveness. However, contrastive learning often involves complex strategies to construct negative samples.
For example, \citet{cao2021cliff} use a combination of multiple methods including entity swapping, content masking and refilling, and low-confidence model generations. 

Our work falls into the data augmentation and contrastive learning category. We adopt LLMs to construct negative samples with more diversity compared to previous strategies that have been predominantly driven by rules and heuristics.

\begin{figure*}[ht]
\centering
\includegraphics[width=0.8\linewidth]{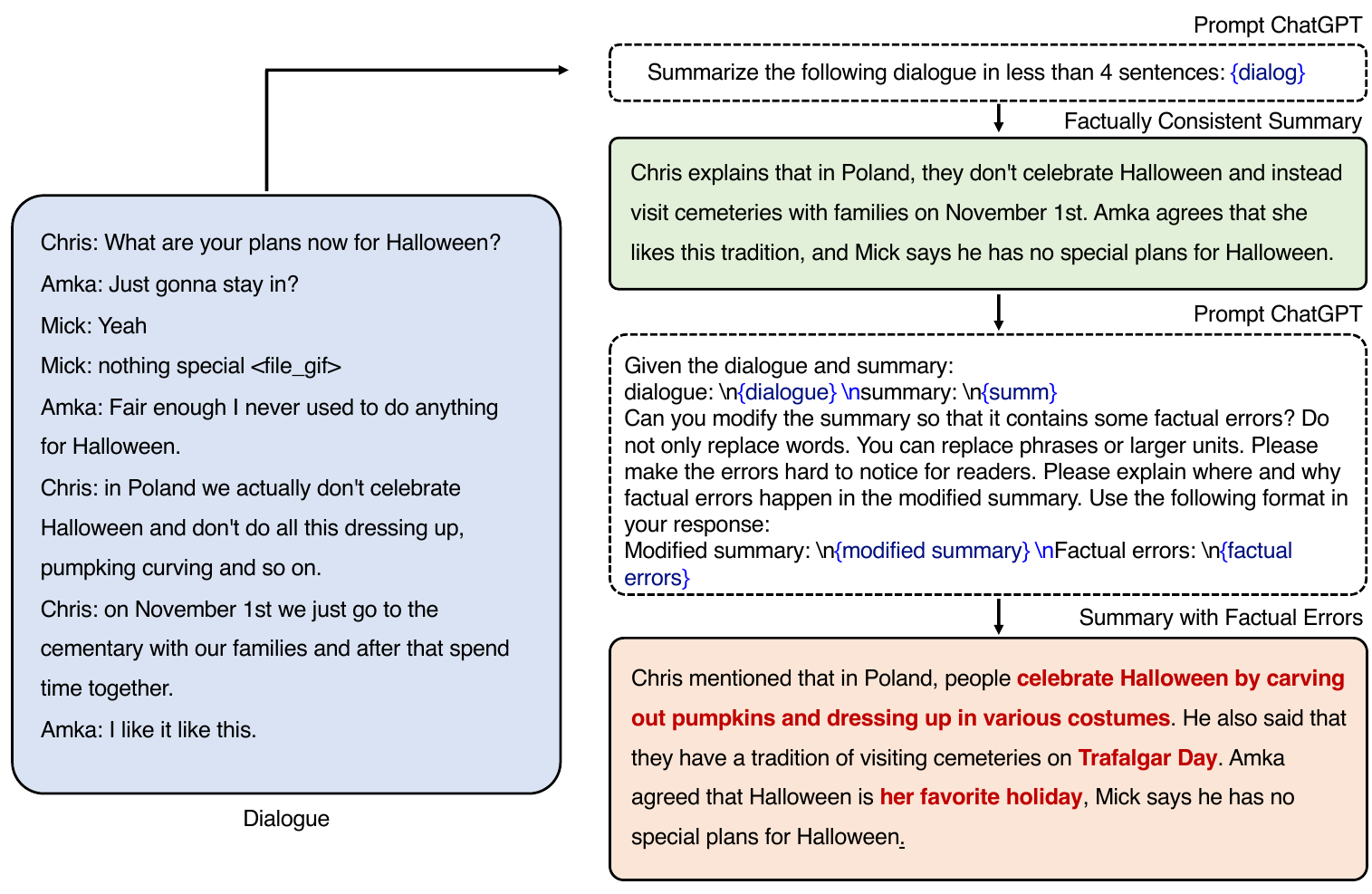}
\caption{To extract symbolic knowledge from the teacher model (ChatGPT) for contrastive learning, we first prompt ChatGPT to generate a factually consistent summary, then use another prompt to instruct ChatGPT to modify the summary into a factually inconsistent version. The \textcolor{red}{contents in red} contain factual errors against the source dialogue.}
\label{fig:prompt_example}
\end{figure*}

\subsection{Symbolic Knowledge Distillation}
Symbolic knowledge distillation~\citep{west2022symbolic} is a conceptual framework originally proposed for constructing common-sense knowledge graphs~\citep{sap2019atomic}. A key advantage of the framework is that it does not require optimizing the student model on the teacher model's output probabilities, which was done in standard knowledge distillation~\citep{hinton2015distilling}. Instead, it extracts symbolic knowledge (e.g., text) from the teacher model to construct a smaller student model. 

Symbolic knowledge distillation has been used to construct better summarization models in different ways, motivated by the high-quality summaries generated by zero-shot and few-shot LLMs~\cite{zhang2023benchmarking}, which are even preferred over human-written summaries. For example, ~\citet{sclar2022referee} construct reference-free sentence summarization models with better controllability on the compression ratio, while ~\citet{song2023enhancing} enhance summary abstractiveness via calibrated distillation. ~\citet{liu2023learning} use LLMs not only as a data augmenter to generate ``quasi-references'', but also as a summary evaluator to provide additional training signals. ~\citet{jiang2024trisum} distill LLM's summarization capability by generating multiple aspect-triple rationales and summaries, then utilize curriculum learning to train student models. 

Our method differs from these studies by incorporating a stage that leverages both positive and negative summaries through contrastive learning to enhance the factual consistency of student models, while the studies above only consider positive examples.

\section{Methodology}

Given a dialogue $D$ (aka ``source documents'' in  document summarization studies), we aim to generate a summary $S$ using a summarization model $g$ that captures the main ideas of $D$. We specifically encourage $S$ to be factually consistent with $D$, i.e., only including information directly found in $D$ and not any information against the facts in $D$.

To construct more factually consistent and cost-effective dialogue summarization models, we first extract symbolic knowledge (i.e., augmented summaries) from a teacher model (ChatGPT), then use sequence-level knowledge distillation and contrastive learning to exploit the knowledge. An overview of our framework is shown in Figure~\ref{fig:outline}.

\subsection{Extracting Symbolic Knowledge}\label{data_creation}
We use ChatGPT (\textit{gpt-3.5-turbo}) to generate positive summaries which are supposed to be factually consistent with the source dialogue $D$, and negative summaries that contain factual errors against $D$. Specifically, we first prompt ChatGPT to generate $k$ {($k = 3$)} positive summaries for a dialogue, then we prompt it again to modify each positive summary into a negative one by modifying snippets of the summary (so we also have $k$ negative summaries). An example is shown in Figure~\ref{fig:prompt_example}. We find that the quality of negative summaries improve when we explicitly prompt ChatGPT to explain the factual errors\footnote{The average factual consistency (AlignScore) for 200 random positive summaries in the training set from the teacher model is 0.90 for SAMSum and 0.92 for DialogSum, indicating that positive summaries are mostly factually consistent. More details in Appendix~\ref{appendix:quality_of_chatgpt_summs}.}.

\subsection{Utilising Symbolic Knowledge}
The standard method to train summarization models is Maximum Likelihood Estimation (\MLE). Specifically, given a single reference summary $R^*$, the summarization model $g$ is encouraged to give the $i$-th token of $R^*$ the maximum probability among all tokens in the vocabulary, based on the prefix string of the current token. The loss function, cross entropy, is defined as follows:
\begin{equation}\label{eq:cross_entropy}
\begin{aligned}
    l_{mle} &= -\log_{}(R^*|D) \\
    &= -\sum_{i=1}^{n}{\log{P_g(R^*_i|D, R^*_{<i})}}
\end{aligned}
\end{equation}
Here, $R^*_i$ is the $i$-{th} token in $R^*$; $R^*_{<i}$ represents the tokens preceding $R^*_i$; and $P_g$ is the probability distribution of the summarization model. Since there is only one reference summary, the loss function encourages the model to approximate the point mass distribution defined by the single reference~\cite{liu2023learning}. As the loss function is defined at the word level in an autoregressive manner, it does not explicitly facilitate the factual consistency of the generated summary, which requires signals at semantic level and sequence level. 

\subsubsection{Sequence-level Distillation}
Given that a large teacher model may generate more factually consistent summaries than the smaller student models, we employ Sequence-level Knowledge Distillation (\SeqDistill) ~\citep{kim2016sequence}. This approach involves generating multiple quasi-summaries from the teacher model, which are then utilized as targets for fine-tuning the student models using cross-entropy loss. Given a set of positive summaries $\mathcal{P^*}$ generated by the teacher model, and the original human-written reference summary $R^*$, the loss function is as follows:
\begin{equation*}
\resizebox{0.9\columnwidth}{!}{$
    l_{s} = -\frac{1}{|\mathcal{P^*} \cup \{R^*\}|} \sum\limits_{R \in \mathcal{P^*} \cup \{R^*\}}{\log{P_g(R|D)}}
$}
\end{equation*}

The primary distinction between \SeqDistill and Maximum Likelihood Estimation (MLE) lies in their method of distribution approximation. \SeqDistill aims to approximate the teacher model's distribution, favoring multiple factually consistent summaries via a sampling-based method. Conversely, MLE approximates a point-mass distribution, where a single reference summary is given all the probability mass.

\subsubsection{Contrastive Learning}
We further incorporate two types of contrastive learning methods to boost the factual consistency of summarization models by incorporating negative summaries on top of \SeqDistill. 

Let $\mathcal{P}$ be a set of \textit{positive summaries} that are factually consistent with the source dialogue $D$, $\mathcal{N}$ be a set of \textit{negative summaries} that contain factual errors against $D$, and $R$ be the target for cross entropy loss. A training instance with contrastive learning is a tuple $(D, R, \mathcal{P},  \mathcal{N})$. The loss function for a single training instance is defined as:
\begin{equation}
l = l_{mle} + \alpha \cdot l_c
\end{equation} 
where $l_c$ is the contrastive loss, $\alpha \in [0, 1]$ is a hyperparameter to balance the two loss terms. 
Intuitively, $l_c$ serves as a regularization term that shapes the distribution of the summarization model to favor factually consistent summaries. We employ two contrastive objectives, \MarginContrast and \PairContrast, which differentiate between positive and negative summaries at the sequence and latent representation level, respectively.

\vspace{5pt}
\noindent \textbf{\MarginContrast} aims to pull apart the positive summaries and negative summaries by enforcing a gap between sequence-level scores. Specifically, we aim to achieve higher scores for even the \textit{worst positive summaries} than those of the \textit{best negative summaries}, with the following loss: 
\begin{equation}
    l_c = \max\{{0, \theta + \max\{S(\mathcal{N})\} - \min\{S(\mathcal{P})\}}\}
\end{equation}
Here, $\theta$ is the target score threshold, and $S(\mathcal{\cdot})$ is a scoring function. Inspired by BARTScore~\citep{yuan2021bartscore}, we define the scoring function $S(\mathcal{\cdot})$ for a summary $X$ using the summarization model $g$ as the length-normalized log-likelihood of all tokens:
\begin{equation}
    S(X) = \frac{1}{m}\sum_{i=1}^{m}\log{P_g}(x_i|D, X_{<i})
\end{equation}
Here, $m$ represents the number of tokens in $X$; $x_i$ is the $i$-th token; and $X_{<i}$ are the preceding tokens. Normalizing by $m$ eliminates the impact of length on the evaluation of factual consistency.

\vspace{5pt}
\noindent \textbf{\PairContrast} differentiates positive from negative summaries by minimizing the similarities between their latent representations, while simultaneously maximizing the similarities among positive pairs.
Let $r_i$, $r_j$, and $r_k$ be summaries from either $\mathcal{P}$ or $\mathcal{N}$. We use $\mathbf{h_i}$ $\mathbf{h_j}$, and $\mathbf{h_k}$ to denote the vector-form representations of these summaries.  
The contrastive loss $l_c$ is defined in accordance with the fomulation provided by~\citet{cao2021cliff} as follows:
\begin{equation}\label{eq:cliffcontrast}
    l_{c} = -\frac{1}{{|\mathcal{P}| \choose 2}}\sum\limits_{\substack{r_i, r_j \in \mathcal{P} \\ r_i \neq r_j}}\log\frac{\exp({\text{s}(\mathbf{h_i}, \mathbf{h_j})/\tau})}{\sum\limits_{\substack{r_k \in \mathcal{P} \cup \mathcal{N} \\ r_k \neq r_i}}{\exp(\text{s}(\mathbf{h_i}, \mathbf{h_k})/\tau)}}
\end{equation}
Here, $\text{s}$ is the \textit{cosine} function; and $\tau$ is a temperature parameter ($\tau$=1 {in our experiments}). 
We follow~\citet{cao2021cliff} to obtain the vector representations of the summaries by applying an MLP projection to the averaged last-layer outputs from the decoder for all tokens. 

To summarize, \MarginContrast uses summary log-likelihood estimated by the summarization model directly, while \PairContrast relies on the internal representation of summary words.

\section{Experiment Setup}
\subsection{Datasets}\label{datasets}
We adopt two popular dialogue summarization datasets: SAMSum~\citep{gliwa2019samsum} and DialogSum~\citep{chen2021dialogsum}. SAMSum is a collection of messenger-like conversations, while DialogSum contains daily conversations in a more real-life setting. In both datasets, there is one human-written reference summary for each conversation in the training split. Table~\ref{tab:samsum_dialogsum_statistics} shows the statistics of the two datasets.

\begin{table}[t]
\setlength{\tabcolsep}{1pt}
\centering
\begin{tabular}{@{}lcccccc@{}}
\toprule
\small{\textbf{Dataset}}   & \small{\textbf{\#Train}} & \small{\textbf{\#Dev}} & \small{\textbf{\#Test}} & $\frac{\textbf{\#Speakers}}{\textbf{\#dial.}}$ & $\frac{\textbf{\#Turns}}{\textbf{\#dial.}}$ & $\frac{\textbf{\#Tokens}}{\textbf{dial.}}$ \\ \midrule
\small{SAMSum}    & 14,732   & 818   & 819    & 2.39       & 9.5              & 94                \\
\small{DialogSum} & 12,460   & 500   & 500    & 2.01       & 11.1             & 131                    \\ \bottomrule
\end{tabular}%
\caption{Dataset statistics. $\textbf{\#Train}$, $\textbf{\#Dev}$ and $\textbf{\#Test}$ refer to the numbers of dialogue-summary pairs (one summary per dialogue) in the training, development, and testing subsets. $\frac{\textbf{\#Speakers}}{\textbf{\#dial.}}$, $\frac{\textbf{\#Turns}}{\textbf{\#dial.}}$, and $\frac{\textbf{\#Tokens}}{\textbf{dial.}}$ refer to the average numbers of speakers, turns, and tokens in each dialogue.}
\label{tab:samsum_dialogsum_statistics}
\end{table}

\subsection{Student Models}\label{backbone_models}
We choose BART~\citep{lewis2020bart}, PEGASUS~\citep{zhang2020pegasus} and Flan-T5~\citep{chung2024scaling} as the student models, which have consistently demonstrated state-of-the-art performance in automatic text summarization~\citep{zhao2022calibrating, liu2021simcls, chung2024scaling}. Specifically, we use \textit{facebook/bart-large}, \textit{google/pegasus-large}, \textit{google/flan-t5-large} as initial checkpoints. The number of learnable parameters for these models are 406 million, 568 million and 770 million, respectively, which are much smaller than that of the teacher model.

\subsection{Baseline Models}
\textbf{\FACTPEGASUS}~\citep{wan2022factpegasus}: an abstractive text summarization model for news summarization. It enhances factual consistency through several strategies: (1) factuality-oriented pre-training, (2) reference summary correction that addresses potential factual errors in reference summaries, (3) contrastive learning to boost the model's ability to differentiate between positive and negative summaries, where the negative summaries are constructed by rule-based entity swapping, (4) pre-training task simulation during fine-tuning that minimizes the gap between the pre-training and fine-tuning phases. We used their pre-trained model and code to fine-tune on our datasets.\footnote{\url{https://github.com/meetdavidwan/factpegasus}}

\textbf{\SWING}~\citep{huang2023swing}: an abstractive dialogue summarization model that achieves state-of-the-art factual consistency and coverage on \SAMSUM and \DIALOGSUM. It leverages an uncovered loss to boost information coverage, and a contrastive loss to enhance factual consistency. We use their model generations directly.\footnote{\url{https://github.com/amazon-science/AWS-SWING}}

We also include the original human-written reference summaries (\HUMANREF) to assess the relative quality compared to our method.

\begin{table*}[th]
\setlength{\tabcolsep}{4pt}
\centering
\begin{adjustbox}{max width=\linewidth}
\begin{tabular}{rc@{\;}cc@{\;}c@{\;}cc@{\;}c@{\;\;\;\;\;\;}c@{\;}cc@{\;}c@{\;}cc@{\;}c}
\toprule
\multicolumn{1}{l}{}     & \multicolumn{7}{c}{\textbf{SAMSum}}                                                                      & \multicolumn{7}{c}{\textbf{DialogSum}}                                                             \\ \cmidrule(l){2-15} 
\multicolumn{1}{c}{}               & \multicolumn{2}{c}{\textbf{Const}} & \multicolumn{3}{c}{\textbf{UniEval}} & \multicolumn{2}{l}{\textbf{ROUGE}}       & \multicolumn{2}{c}{\textbf{Const}} & \multicolumn{3}{c}{\textbf{UniEval}} & \multicolumn{2}{l}{\textbf{ROUGE}} \\ \midrule
\multicolumn{1}{r}{\textbf{Model}}          & $\mathbf{S_A}$    & $\mathbf{S_G}$   & \textbf{Coh}   & \textbf{Flu}  & \textbf{Rel}  & \textbf{R1} & \textbf{R2}   & $\mathbf{S_A}$           & $\mathbf{S_G}$           & \textbf{Coh}     & \textbf{Flu}     & \textbf{Rel}     & \textbf{R1}          & \textbf{R2}          \\ \midrule

\multicolumn{1}{r}{\HUMANREF}    & 0.80            & 4.80            & 0.92    & 0.93    & 0.97    & 1.00    & 1.00    & 0.82           & 4.84           & 0.94    & 0.92    & 0.98    & 1.00           & 1.00           \\ \midrule

\multicolumn{15}{c}{Baselines}                                                                                                                                                                                                     \\ \midrule
\multicolumn{1}{r}{\FACTPEGASUS} & 0.63           & 3.08           & 0.87    & 0.90     & 0.73    & 0.45 & 0.20  & 0.67           & 3.44           & 0.88    & 0.87    & 0.77    & 0.49        & 0.24        \\

\multicolumn{1}{r}{\SWING}         & 0.82           & 4.38           & 0.93    & 0.93    & 0.84    & 0.52 & 0.28 & 0.83           & 4.54           & 0.95    & 0.93    & 0.90     & 0.53        & 0.29        \\ \midrule
\multicolumn{15}{c}{MLE}                                                                                                                                                                                                           \\ \midrule
\multicolumn{1}{r}{BART}           & 0.82           & 4.27           & 0.92    & 0.93    & 0.84    & \textbf{0.52} & \textbf{0.28} & 0.80           & 4.22           & 0.94    & 0.93    & 0.88    & \textbf{0.53}        & \textbf{0.28}        \\
\multicolumn{1}{r}{PEGASUS}        & 0.81           & 4.12           & 0.93    & 0.94    & 0.84    & \textbf{0.50} & \textbf{0.26} & 0.83           & 4.44           & 0.96    & 0.93    & 0.90     & \textbf{0.52}        & \textbf{0.28}        \\
\multicolumn{1}{r}{Flan-T5}             & 0.82           & 4.34           & 0.93    & 0.93    & 0.84    & \textbf{0.52} & \textbf{0.28} & 0.84           & 4.65           & \textbf{0.96}    & \textbf{0.93}    & \textbf{0.91}    & \textbf{0.54}    & \textbf{0.29}    \\ \midrule
\multicolumn{15}{c}{\SeqDistill (Our Method)}                                                                                                                                                                                       \\ \midrule
\multicolumn{1}{r}{BART}           & 0.87           & 4.41           & 0.96    & \textbf{0.94}   & 0.89    & 0.36 & 0.14 & \textbf{0.93}           & \textbf{4.81}           & \textbf{0.98}    & 0.93    & \textbf{0.93}    & 0.29        & 0.13        \\
\multicolumn{1}{r}{PEGASUS}        & \textbf{0.89}           & \textbf{4.52}           & 0.95    & \textbf{0.94}    & \textbf{0.89}    & 0.39 & 0.17 & 0.90            & \textbf{4.73}           & \textbf{0.97}    & 0.93    & \textbf{0.91}    & 0.42        & 0.22        \\
\multicolumn{1}{r}{Flan-T5}             & 0.88           & 4.51           & 0.94    & 0.93    & 0.87    & 0.40  &0.17 & 0.91           & 4.80            & \textbf{0.96}    & \textbf{0.93}    & 0.90     & 0.32        & 0.15        \\ \midrule
\multicolumn{15}{c}{\MarginContrast (Our Method)}                                                                                                                                                                                   \\ \midrule
\multicolumn{1}{r}{BART}           & 0.89           & \textbf{4.73}           & 0.97    & \textbf{0.94}    & 0.90     & 0.40  & 0.18 & \textbf{0.93}           & 4.72           & \textbf{0.98}    & \textbf{0.94}    & \textbf{0.93}    & 0.31        & 0.15        \\
\multicolumn{1}{r}{PEGASUS}        & 0.87           & 4.08           & 0.92    & \textbf{0.94}    & 0.84    & 0.38 & 0.17 & 0.89           & 4.31           & 0.95    & 0.93    & 0.88    & 0.34        & 0.17        \\
\multicolumn{1}{r}{Flan-T5}             & 0.90            & 4.69           & 0.95    & \textbf{0.94}    & 0.88    & 0.42 & 0.20  & 0.91           & 4.76           & 0.95    & \textbf{0.93}    & 0.90     & 0.37        & 0.19        \\ \midrule
\multicolumn{15}{c}{\PairContrast (Our Method)}                                                                                                                                                                                     \\ \midrule
\multicolumn{1}{r}{BART}           & \textbf{0.91}           & 4.69           & \textbf{0.98}    & \textbf{0.94}    & \textbf{0.92}    & 0.37 & 0.15 & \textbf{0.93}           & 4.80            & \textbf{0.98}    & 0.93    & \textbf{0.93}    & 0.30         & 0.14        \\
\multicolumn{1}{r}{PEGASUS}        & \textbf{0.89}          & 4.47           & \textbf{0.96}    & \textbf{0.94}    & \textbf{0.89}    & 0.38 & 0.16 & \textbf{0.91}           & 4.62           & 0.96    & \textbf{0.94}    & \textbf{0.91}    & 0.36        & 0.18        \\
\multicolumn{1}{r}{Flan-T5}             & \textbf{0.91}           & \textbf{4.74}           & \textbf{0.96}    & \textbf{0.94}    & \textbf{0.90}     & 0.38 & 0.16 & \textbf{0.93}           & \textbf{4.86}           & \textbf{0.96}    & \textbf{0.93}    & 0.89    & 0.37        & 0.19        \\ \bottomrule
\end{tabular}
\end{adjustbox}
\caption{Comparing different models and training strategies on Consistency (Const), Coherence (Coh), Fluency (Flu), Relevance (Rel) and ROUGE. We use two automatic factual consistency metrics, AlignScore ($S_A$) and G-Eval ($S_G$). Coherence, Fluency and Relevance are obtained from UniEval. R1 and R2 represent the F1 score of ROUGE 1 and ROUGE 2, respectively. We show the highest score(s) in all columns for the same model (e.g., BART) across \{\MLE, \SeqDistill, \MarginContrast, \textsc{PairContrast}\} in \textbf{bold} to show the most effective training strategy.}
\label{tab:main_results}
\end{table*}

\subsection{Evaluation Metrics}
We selected multiple reference-free evaluation metrics, recognizing that our methods may produce high-quality summaries that diverge from human-written references. This divergence could lead to underrating by reference-based metrics. To assess factual consistency, we employed two state-of-the-art (SOTA) automatic metrics: an LLM-based metric, \GEval~\citep{liu-etal-2023-g}, and a non-LLM-based metric, \AlignScore~\citep{Zha2023AlignScoreEF}~\footnote{Our meta-evaluation on multiple dialogue summarization datasets show that AlignScore and G-Eval exhibit high correlation (0.4-0.7) with human evaluation results. More details in Appendix~\ref{appendix:factual_metric_metaeval}.}. This approach mitigates the potential bias of favoring LLM-generated summaries inherent in LLM-based metrics~\citep{liu-etal-2023-g}.
Additionally, we used \UniEval~\citep{Zhong2022TowardsAU} to evaluate Coherence, Fluency, and Relevance. We also utilized the standard n-gram matching-based metric,
ROUGE~\citep{lin2004rouge}, primarily as a sanity check for models trained using \MLE.

\subsection{Other Experimental Details}
For \MarginContrast and \PairContrast, we merge the human-written reference $R^*$ and positive summaries $\mathcal{P^*}$ generated by the teacher model as the positive set $\mathcal{P}' = \{R^*\} \cup \mathcal{P^*}$. For each training sample, we select one element $R \in \mathcal{P}'$ as the target for cross-entropy loss and use the rest as $\mathcal{P}$ for contrastive loss. All models are fine-tuned for 15,000 steps and evaluated at every 500 steps. The best checkpoint is selected according to AlignScore on the development set. We provide more implementation details in Appendix~\ref{appendix:implementation_details}.

\section{Results and Discussions}\label{sec:experiments}

\subsection{The Effectiveness of Symbolic Knowledge Distillation and Contrastive Learning}

We compare the performance of our methods (\SeqDistill, \MarginContrast and \PairContrast) and the baseline models on various quality dimensions, with a focus on factual consistency. From the results in Table~\ref{tab:main_results}, we make the following observations:
\begin{itemize}
    \item Our distillation methods improve factual consistency (compared to baseline models and MLE methods) without sacrificing in other quality dimensions (i.e., Coherence, Fluency and Relevance).
    \item Our distillation methods consistently enhance the factual consistency of all pretrained models (BART, PEGASUS and Flan-T5). \PairContrast is generally the most effective method, although there is some performance variation depending on the dataset and pretrained model.
    \item \SeqDistill and two contrastive learning methods result in significantly lower Rouge scores compared to MLE. However, it only tells us that there are fewer word overlaps between model generated summaries and human-written references rather than an actual quality decline. We will revisit this again with a case study in section~\ref{sec:case_study}.
    \item Flan-T5 in most cases generate more factually consistent summaries than BART and PEGASUS across different settings (\MLE, \SeqDistill, \MarginContrast, \PairContrast).
    \item  Flan-T5 with \PairContrast is the best summarization model overall, and it achieves comparable or sometimes better factual consistency, coherence and fluency than \HUMANREF according to $S_A$, $S_G$ and \UniEval. 
\end{itemize}

\subsection{The Effect of Human-written References}
Observing that the best-performing student model demonstrates promising results, we further explore the impact of human-written references and seek to address the question: \textit{Is it possible to construct dialogue summarization models without human-written references?}

Table~\ref{tab:ori_ref} displays the performance of \textit{flan-t5-large} trained using \PairContrast with various numbers of randomly sampled dialogues from the SAMSum training set. The quality scores on SAMSum test set across all dimensions are similar, whether original human-written reference summaries are employed ($R^=Y$) or not ($R^=N$), for all dataset sizes. These findings suggest the feasibility of developing robust summarization models using unlabeled datasets.

\begin{table}[t]
\centering
\begin{tabular}{@{}rccccc@{}}
\toprule
\#Dialog & $R^*$ & Const & Coh  & Flu  & Rel \\ \midrule
300   & N & 0.89 & 0.96 & 0.93 & 0.88 \\
300   & Y & 0.88 & 0.94 & 0.91 & 0.83 \\
1000  & N & 0.89 & 0.94 & 0.92 & 0.86 \\
1000  & Y & 0.89 & 0.95 & 0.93 & 0.86 \\
3000  & N & 0.90 & 0.96 & 0.94 & 0.89 \\
3000  & Y & 0.90 & 0.95 & 0.93 & 0.88 \\
9000  & N & 0.91 & 0.96 & 0.93 & 0.88 \\
9000  & Y & 0.90 & 0.96 & 0.94 & 0.89 \\
13000 & N & 0.91 & 0.96 & 0.94 & 0.89 \\
13000 & Y & 0.91 & 0.96 & 0.94 & 0.89 \\ 
\bottomrule
\end{tabular}
\caption{Comparing the performance of \textit{flan-t5-large} with \PairContrast on SAMSum, with ($R^* = Y$) or without ($R^*=N$) human-written references. $k=3$ for all settings. The four quality dimensions are factual consistency (Const), coherence (Coh), fluency (Flu) and relevance (Rel). Factual consistency is obtained from AlignScore.}
\label{tab:ori_ref}
\end{table}

\vspace{6pt}
\subsection{The Effect of the Number of Contrastive Pairs}

\begin{table}[t]
\centering
\begin{tabular}{@{}ccc@{}}
\toprule
\#Dialog & $k$ & Consistency \\ \midrule
1000     & 3      & 0.893          \\
3000     & 1      & 0.898          \\
3000     & 2      & 0.905          \\
3000     & 3      & 0.902          \\
9000     & 1      & 0.902          \\
9000     & 2      & 0.904          \\
9000     & 3      & 0.913          \\ \bottomrule
\end{tabular}
\caption{Factual consistency (AlignScore) of \textit{flan-t5-large} trained with \PairContrast on varying numbers of dialogues (\#Dialog) and contrastive pairs per dialogue ($k$).}
\label{tab:consistency_npos}
\end{table}

\begin{figure*}[th]
\centering
\includegraphics[width=\linewidth]{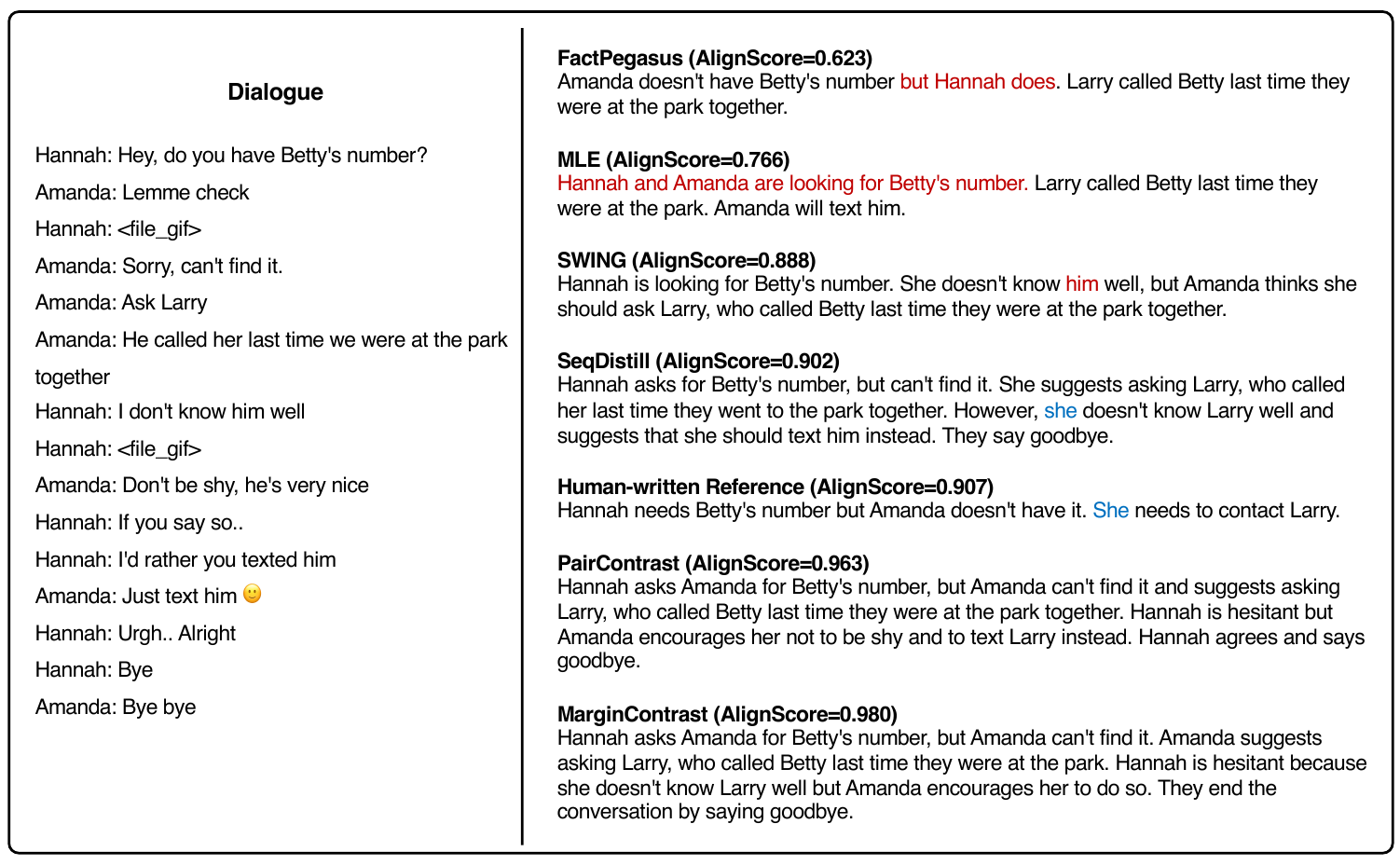}
\caption{An example dialogue from \SAMSUM~\citep{gliwa2019samsum} with summaries generated by BART~\citep{lewis2020bart} trained with different strategies (\MLE, \SeqDistill, \MarginContrast, \PairContrast). Baseline models (FactPEGASUS, SWING) and human-written reference are included for comparison. Contents that are \textcolor[HTML]{d0413b}{inconsistent} with the input dialogue are shown in \textcolor[HTML]{d0413b}{red}. \textcolor[HTML]{1079c3}{Ambiguous} contents are shown in \textcolor[HTML]{1079c3}{blue}. }
\label{fig:case_study}
\end{figure*}

Table~\ref{tab:consistency_npos} further shows the performance of \textit{flan-t5-large} trained on different numbers of dialogues and contrastive pairs. We see that when the number of dialogues (i.e., \#Dialog) is fixed, the model in general generates slightly more consistent summaries as 
$k$ grows. On the other hand, there is no significant difference when we vary the number of contrastive pairs as long as the total number of  training instances (i.e., \#Dialog $\times$ $k$) is fixed. For example, when the total number of training instances is 9,000, (\#Dialog=3000, $k$=3) yields the same result as (\#Dialog=9000, $k$=1) does.

\vspace{6pt}
\subsection{Case Study}\label{sec:case_study}

Figure~\ref{fig:case_study} presents an example dialogue along with summaries generated by different models, sorted by AlignScore~\citep{Zha2023AlignScoreEF} in ascending order. The summaries from \FACTPEGASUS, \MLE, and \SWING include factual errors unsupported by the dialogue. Specifically, \FACTPEGASUS incorrectly asserts ``but Hannah does'' when in fact, Hannah does not have Betty's number. \MLE inaccurately claims that ``Hannah and Amanda are looking for Betty's number'', though only Hannah is searching. In \SWING's summary, ``him'' appears before the referent ``Larry''. For \SeqDistill and Human-written reference, the pronouns ``she'' are ambiguous as there are multiple possible referent in previous context. Unlike these, summaries from \PairContrast and \MarginContrast do not contain ambiguous references. Notably, our methods (\SeqDistill, \PairContrast and \MarginContrast) tend to produce longer summaries compared to the much more succinct human-written references, hence we see a substantially lower ROUGE scores for them (Table~\ref{tab:main_results}).

\section{Conclusion}
We investigated distilling LLM's symbolic knowledge (in the form of generated summaries) to enhance the factual consistency of smaller models for dialogue summarization. Our experiments with BART, PEGASUS, and Flan-T5 on the SAMSum and DialogSum datasets reveal that: (1) symbolic knowledge distillation enables the creation of more compact summarization models that surpass strong baselines which use complex data augmentation strategies; and (2) our best-performing student model, Flan-T5 with \PairContrast, produces summaries that are potentially better --- in terms of factual consistency, coherence and fluency --- than human-written references.

\section{Limitations}
The experiments in this paper are conducted on short daily dialogues. The findings may not generalize to other dialogue scenarios such as academic meetings and television interviews.

We use automatic evaluation metrics to assess the quality of model-generated summaries, which may not fully reflect human preferences.

\section{Ethics Statement}
This study is conducted under the guidance of the ACL code of Ethics.

\section*{Acknowledgements}
This research was supported by The University of Melbourne’s Research Computing Services and the Petascale Campus Initiative.

\bibliography{anthology,custom}

\appendix

\section{Appendix}\label{sec:appendix}

\subsection{Potential Risks}
The summaries generated by ChatGPT may contain social biases, which require further investigation in real applications.

\subsection{The Statistics and Quality of ChatGPT Summaries}\label{appendix:quality_of_chatgpt_summs}
We generated 3 positive and 3 negative summaries for 13,000 dialogues from the training split of SAMSum and 11,000 dialogues from the training split of DialogSum. For each dialogue, we made 6 API calls (3 for positive and 3 for negative) separately.

Table~\ref{tab:chatgpt_possumm_quality} shows the quality of 200 randomly sampled positive summaries generated by the teacher model \textit{gpt-3.5-turbo}, validating that these summaries are mostly factually consistent, with high coherence, fluency and relevance as well.

\begin{table}[th]
\centering
\begin{tabular}{@{}rcccc@{}}
\toprule
Dataset   & Const & Coh  & Flu  & Rel  \\ \midrule
SAMSum    & 0.90  & 0.97 & 0.94 & 0.91 \\
DialogSum & 0.92  & 0.97 & 0.94 & 0.94 \\ \bottomrule
\end{tabular}
\caption{The factual consistency (Const), coherence (Coh), fluency (Flu) and relevance (Rel) for 200 randomly sampled positive summaries, generated by \textit{gpt-3.5-turbo}, in the training set of SAMSum and DialogSum. Factual consistency is obtained from AlignScore~\citep{Zha2023AlignScoreEF}. Coherence, fluency, and relevance are obtained from UniEval~\citep{zhong2022towards}. }
\label{tab:chatgpt_possumm_quality}
\end{table}

\subsection{Meta-evaluation of Factual Consistency Evaluation Metrics}\label{appendix:factual_metric_metaeval}

We conducted a meta-evaluation of various automatic factual consistency metrics across three datasets: DiaSummFact~\citep{zhu2023annotating}, FacEval~\citep{wang2022analyzing}, and GO FIGURE~\citep{gabriel2021go}. For the GO FIGURE dataset, we specifically utilized the subset derived from SAMSum~\citep{gliwa2019samsum}. In the case of DiaSummFact, we conducted evaluations at both the sentence level (DiaSummFact$^*$) and summary level (DiaSummFact'). For the sentence-level evaluation, we excluded sentences whose labels include ``Link Error'' or ``Coreference Error''. All labels across the datasets were converted into a binary format: if any category of factual error is present, the label is marked as ``factually inconsistent''; otherwise, it is marked as ``factually consistent''. The number of (dialogue, output) pairs in each dataset, where the output is either a sentence for sentence-level evaluation or a summary for summary-level evaluation, is presented in Table~\ref{tab:meta_eval_statistics}. Spearman and Pearson correlations are shown in Table~\ref{tab:meta_eval_spearman_corr} and Table~\ref{tab:meta_eval_pearson_corr}.

Results show that both AlignScore and G-Eval exhibit high correlation with human annotations in most cases, except AlignScore on FacEval, which requires further investigation in future works. UniEval shows unsatisfactory correlation with human annotations on factual consistency, thus we only use AlignScore and G-Eval (\textit{gpt-4}) for factual consistency evaluation.

\begin{table}[t]
\centering
\begin{tabular}{@{}rc@{}}
\toprule
            & N \\ \midrule
DiaSummFact$^*$ & 475       \\
DiaSummFact' & 1240      \\
FacEval     & 750       \\
GO FIGURE   & 250       \\ \bottomrule
\end{tabular}
\caption{The number of (dialogue, output) pairs ($N$) in the datasets for our meta-evaluation.}
\label{tab:meta_eval_statistics}
\end{table}

\begin{table}[t]
\centering
\begin{tabular}{@{}rccc@{}}
\toprule
Metric            & AlignScore & G-Eval & UniEval \\ \midrule
DiaSummFact$^*$  & 0.52       & 0.53   & 0.22    \\
DiaSummFact' & 0.48       & 0.60   & 0.15    \\
FacEval           & 0.11       & 0.54   & 0.01    \\
GoFigure          & 0.43       & 0.60   & 0.23    \\ \bottomrule
\end{tabular}
\caption{Spearman correlation between automatic factual consistency evaluation metrics and human evaluation (binary).}
\label{tab:meta_eval_spearman_corr}
\end{table}

\begin{table}[t]
\centering
\begin{tabular}{@{}llll@{}}
\toprule
Metric            & AlignScore & G-Eval & UniEval \\ \midrule
DiaSummFact$^*$  & 0.49       & 0.54   & 0.17    \\
DiaSummFact' & 0.39       & 0.49   & 0.13    \\
FacEval           & 0.09       & 0.49   & -0.01   \\
GoFigure          & 0.44       & 0.71   & 0.23    \\ \bottomrule
\end{tabular}
\caption{Pearson correlation between automatic factual consistency evaluation metrics and human evaluation (binary).}
\label{tab:meta_eval_pearson_corr}
\end{table}

\subsection{Implementation Details}\label{appendix:implementation_details}
All models were fine-tuned for 15,000 steps with a batch size of 32 (per-device batch size 2/1, with gradient accumulation 16/32), evaluated every 500 steps (with model generations on development set) on an NVIDIA A100 GPU with 40G/80G memory. Each training task took between 4 to 72 hours, depending on the size of the model.

We searched for the best hyper-parameters of $\alpha \in \{0.5, 1, 2\}$ for \PairContrast, and $\alpha \in \{0.5, 1, 2\}$ and $\theta \in \{15, 30\}$ for \MarginContrast, according to AlignScore~\citep{Zha2023AlignScoreEF} on development set.

The code for \PairContrast was developed based on CLIFF~\footnote{\url{https://github.com/ShuyangCao/cliff_summ/tree/main/models}}.
ROUGE scores are computed using Python package \textbf{evaluate 0.4.0} with default parameters~\footnote{\url{https://pypi.org/project/evaluate/}}.

\subsection{License or Terms}
Our code and data will be released under MIT license.

\subsection{Intended Use of Existing Artifacts}
The SAMSum dataset, as presented in~\citet{gliwa-etal-2019-samsum}, is distributed under the Attribution-NonCommercial-NoDerivatives 4.0 International (CC BY-NC-ND 4.0) license. We offer supplementary details (e.g., model-generated summaries), while preserving the integrity of the original data, comprising dialogues and reference summaries.

\subsection{Artifacts}
The artifacts we release (code, data) are all in English only.

\end{document}